\documentclass[sn-mathphys]{sn-jnl}
\usepackage{amsmath}
\newcommand\norm[1]{\left\lVert#1\right\rVert}

\usepackage{multirow}
\usepackage{algorithm, algpseudocode}
\usepackage{commath}
\usepackage{eucal}
\usepackage{siunitx}
\usepackage{graphicx}
\usepackage{textcomp}
\usepackage{xcolor}
\usepackage[T1]{fontenc}

\jyear{2021}%

\theoremstyle{thmstyleone}%
\theoremstyle{thmstyletwo}%

\theoremstyle{thmstylethree}%

\begin{document}

\title[Optimal Stroke Learning with Policy Gradient Approach for Robotic Table Tennis]{Optimal Stroke Learning with Policy Gradient Approach for Robotic Table Tennis}


\author*[1]{\fnm{Yapeng} \sur{Gao}}\email{yapeng.gao@uni-tuebingen.de}

\author[1]{\fnm{Jonas} \sur{Tebbe}}\email{jonas.tebbe@uni-tuebingen.de}

\author[1]{\fnm{Andreas} \sur{Zell}}\email{andreas.zell@uni-tuebingen.de}

\affil*[1]{\orgdiv{Cognitive Systems}, \orgname{Eberhard Karls University T\"ubingen}, \orgaddress{\street{Geschwister-Scholl-Platz}, \city{T\"ubingen}, \postcode{72074}, \state{Baden-W\"urttemberg}, \country{Germany}}}


\abstract{Learning to play table tennis is a challenging task for robots, as a wide variety of strokes required. Recent advances have shown that deep Reinforcement Learning (RL) is able to successfully learn the optimal actions in a simulated environment. However, the applicability of RL in real scenarios remains limited due to the high exploration effort. In this work, we propose a realistic simulation environment in which multiple models are built for the dynamics of the ball and the kinematics of the robot. Instead of training an end-to-end RL model, a novel policy gradient approach with TD3 backbone is proposed to learn the racket strokes based on the predicted state of the ball at the hitting time. In the experiments, we show that the proposed approach significantly outperforms the existing RL methods in simulation. Furthermore, to cross the domain from simulation to reality, we adopt an efficient retraining method and test it in three real scenarios. The resulting success rate is 98\% and the distance error is around 24.9 cm. The total training time is about 1.5 hours.}




\maketitle

\section{Introduction}\label{sec:intro}

Reinforcement learning (RL)~\cite{bertsekas1995neuro, sutton1999reinforcement} is a general learning paradigm that addresses the problem of how an acting agent can learn an optimal behavioral strategy while interacting with an unknown environment surrounding it~\cite{barto2003RL}. Recently, RL has achieved a variety of successes, most prominently in autonomous driving~\cite{Kendall2018, osinski2020simulation}, gaming~\cite{go2017, OpenAI_dota}, and robotics manipulation~\cite{gu2017deep,kalashnikov2018scalable}. Notably, most RL research is based on the formalism of Markov decision process (MDP) which models the decision-making problem. Specifically, while interacting with the environment, the agent takes actions based on the observed state of the environment and then receives rewards for its actions. Thus, the goal of RL is to maximize the expected value of the cumulative reward in an episode. However, formulating a near-optimal policy for the agent's actions usually requires extensive exploration in the action space. For example, OpenAI Five~\cite{OpenAI_dota} defeated the world champions at an esports (Dota2) game by developing a training system using RL techniques. To fully explore the action space, it trains for 180 days on 256 GPUs and 128,000 CPU cores, based on an average of 180 days worth of self-play. Therefore, training RL models in the context of robotics is particularly challenging, since in such a context it is very difficult to safely collect samples that cover all possible actions in their space.

A common method to address this problem is training RL models with the help of simulations~\cite{koos2010crossing, Cutler2015_robSimu,gao2020robotic, mahjourian2018hierarchical, zhu2018}. Previous work has shown that simulation can be used as a valuable tool for robotics research, as performing robotic skills in simulation is comparatively easier than in the real world~\cite{osinski2020simulation, hanna2021grounded}. The use of simulation greatly facilitates the implementation of RL in robotics by allowing a comprehensive exploration of the robotic action space with less engineering effort than in real world by adjusting the parameters of the simulator. However, policies learned in simulation are often unsuitable for reality due to the reality gap. To bridge the gap, researchers usually incorporate data from the real world for training, or retrain the models in reality. For example, a table tennis robot was trained with RL in a hybrid simulation and real system~\cite{buchler2020learning}. 
The real trajectories of the ball were recorded and replayed in simulation to use the real data as much as possible.

\begin{figure}[ht]
	\centering
	\includegraphics[width=0.6\linewidth]{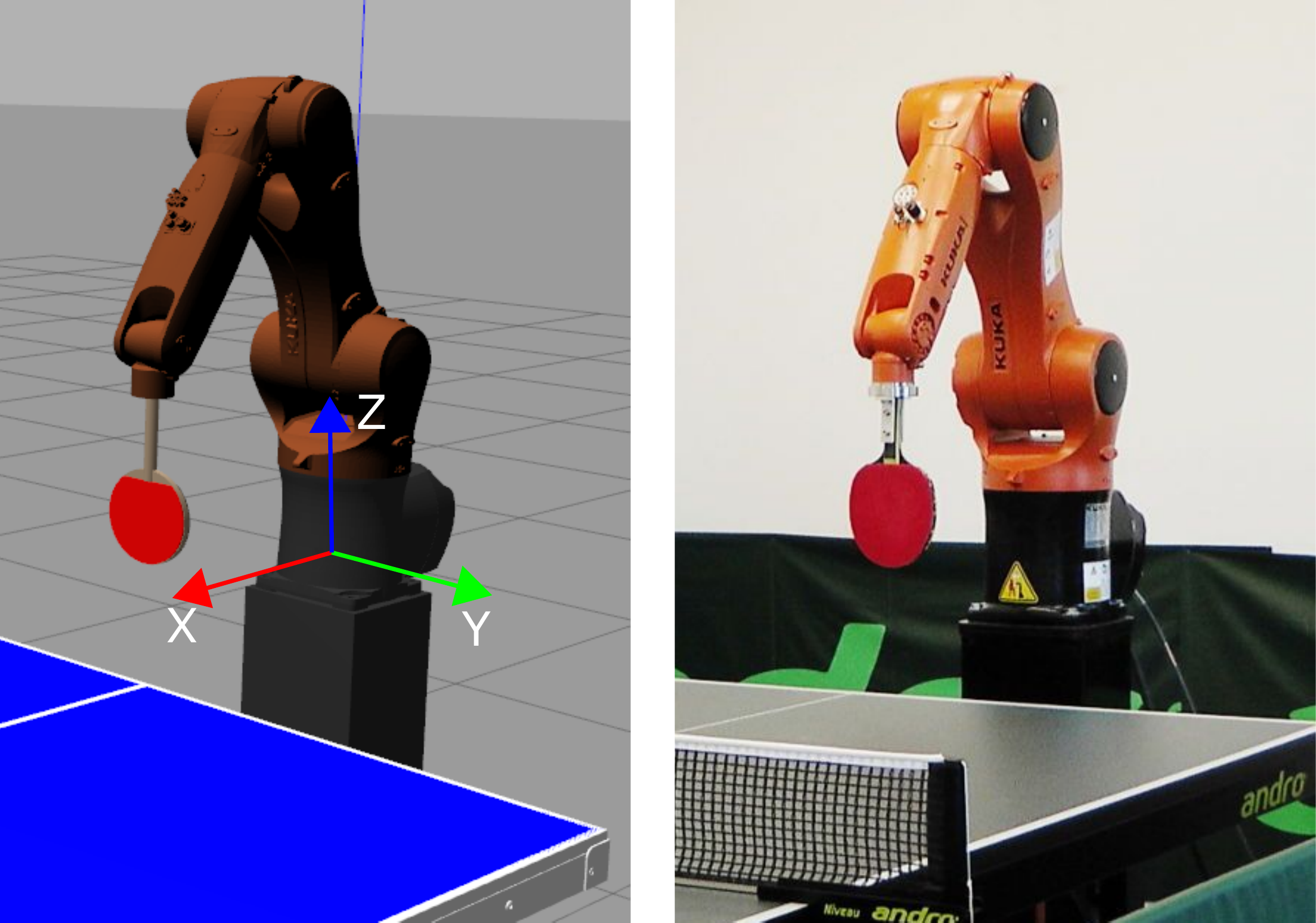}
	\caption{Left: the simulated table tennis robot in a realistic simulation environment. To approximate reality, Gaussian noise is applied to the 3D position of each ball in simulation. The world coordinate system is identical to that of the robot. Right: our table tennis robot with a KUKA KR 6 R900 robot.}
	\label{fig:frame}
\end{figure}

However, one problem in existing RL methods such as Trust Region Policy Optimization (TRPO)~\cite{trpo}, Proximal Policy Optimization
 (PPO)~\cite{ppo}, Deep Deterministic Policy Gradient
 (DDPG)~\cite{ddpg}, Twin Delayed DDPG (TD3)~\cite{td3}, or Soft Actor-Critic
 (SAC)~\cite{sac} is that the fuzzy one-dimensional reward cannot precisely express the interaction with the environment for multi-dimensional actions.
Therefore, inspired by the previous work, we propose a novel approach for optimal stroke learning in robotic table tennis. A 3D $Q$-value function is designed to cope with the corresponding 3D reward vector. Two learning steps, including training in simulation and retraining in reality, can be completed in around 1.5 hours for the balls with a wide variety of spins, speeds, and positions. The main contributions of this work are as follows:

\begin{itemize}	
	\item We design a realistic table tennis robot simulation system for optimal stroke learning with RL, as shown in Fig.~\ref{fig:frame} left. The simulation is based on the Gazebo simulator, the Robot Operating System (ROS), OpenAI Gym~\cite{gym}, and the RL library Spinning Up. Controlled by the Gazebo plugin, the robot and the table tennis ball can publish their states via ROS topics. 	
	\item We decompose the learning strategy into two stages: first, the prediction of the ball's hitting state, and second, the learning of the optimal stroke, which is the focus of this paper. Based on the controllable and applicable actions of the robot, a multidimensional reward function and a $Q$-value model are proposed.
	\item We compare our RL method with others by evaluating them on a dataset of 1000 balls in simulation. An efficient retraining step is used to close the sim-to-real gap. Our models trained in real robots (see Fig. \ref{fig:frame} right) achieve remarkable performance.
	
\end{itemize}

\section{Related work}
\label{sec:related}

\subsection{Simulation for robotic table tennis} 

Just as simulation is being used in various areas of robotics research (e.g., autonomous vehicles~\cite{osinski2020simulation}, industrial robots~\cite{IndustryRobot2019}, and humanoid robots~\cite{humanoidRobot2019}), it is also being widely used for facilitating the development of more intelligent robotic table tennis~\cite{gao2020robotic, Gao_StrokeRec2021}. By using simulation, the robot can explore the action space extensively without worrying about safety during the training steps. Moreover, simulation provides the ability to compare the performance of different approaches in a fair and deterministic environment. For instance,~\cite{mahjourian2018hierarchical} proposed an approach to sample-efficient learning of complex policies in the context of robotic table tennis. The simulation environment was created using PyBullet~\cite{coumans2017pybullet} and connected to a virtual reality setup to capture human actions with instrumented paddles. Similarly, a simulated robotic table tennis system was built using PyBullet~\cite{coumans2017pybullet} to train policies for performing table tennis ball return tasks~\cite{gao2020robotic}. In addition, a simulation environment was developed in MATLAB and used to generate the optimal trajectory for robot table tennis~\cite{kocc2018online}. To make use of robotic drivers and devices,~\cite{Silva2015} developed a flying robot with the Gazebo simulator which can be easily combined with ROS.~\cite{buchler2020learning} presented a hybrid simulation and real (HYSR) training method for muscular robots performing a table tennis task. To teach robots how to play table tennis without using real balls, the historical states of the ball in reality were recorded and replayed in simulation, and then the actions in simulation were applied to real robots. 
The aforementioned works are further evidence of the benefit and importance of using simulations in robotic table tennis research. However, some of these existing simulated table tennis systems have the characteristic that a high effort is required when transferring the trained models from simulation to reality due to the fact that they are not sufficiently realistic. In addition, due to incompatibility with the existing RL libraries that contain advanced RL algorithms, some of these existing simulated table tennis systems are further limited in their generalizability. Therefore, there is an increasing need to develop a more realistic simulation that can better simulate the real situation for advanced robot table tennis research. In this case, we developed a realistic simulation in combination with the Gazebo, ROS, OpenAI Gym and Spinning Up. The dynamics of the ball in simulation is determined based on~\cite{blank2017ball,tebbe2020spin}.
The simulated manipulator in our work is controlled in Cartesian space so that it can be easily replaced by other types of robots, such as flying robots or mobile robots.

\subsection{Reinforcement learning in robotic table tennis} 

Reinforcement learning has been shown to be an excellent method for training robots to learn complex tasks~\cite{hester2013texplore, GuRLrobotsTask}, and therefore could be a solution for robotic table tennis training. Indeed, deep RL has already attracted great interest from researchers and has achieved some success in the field of robotic table tennis~\cite{PeterJointlyTra2016, YangTableRL2021}. For instance,~\cite{gao2020robotic} developed an end-to-end RL algorithm for learning efficient policies to directly control of a simulated table tennis robot in joint space. A multi-modal model-free policy was trained to learn the velocities of each joint at 100 Hz by taking the joint position trajectories and locations of the ball as inputs. The optimal policy was found at about 1 million episodes. In~\cite{buchler2020learning}, a muscular table tennis robot was trained in joint space in a hybrid simulation and reality (HYSR) system. The PPO was used as backbone. To return and smash the ball with a high success rate, a dense reward function was developed depending on the ball position and the robot state. Without using real balls, the robots were trained to play table tennis. 
~\cite{mahjourian2018hierarchical} incorporated stroke learning into a hierarchical control system using an inverse landing model, an analytic racket controller, a forward racket model and a forward landing model. Each model was trained separately to make the learning process easier and more efficient. A striking policy that can hit the ball to the targets with reasonable errors was learned from about 7,000 demonstrated trajectories, and in addition, the agent can learn from about 24,000 strikes in self-play to make optimal use of the human dynamics models for longer play.
In addition, to efficiently learn the optimal stroke, a two-stage approach was adopted in~\cite{zhu2018}. In the first stage, the hitting states of the ball (position and velocity) were determined by an extended Kalman filter (EKF) based predictor. In the second stage, these determined states of the ball were fed as inputs to the DDPG, with the velocities of the racket being the outputs. A reliable performance was achieved within 200,000 simulated episodes. Instead of using DDPG directly,~\cite{tebbe2020sample} proposed an accelerated parametrized-reward gradients approach to learn the velocity and rotation of the racket from the predicted hitting states. The policy was trained with 200 human demonstrations. To keep the explorations safe and avoid collisions, the racket action were restricted to a narrow range, resulting in the network being trained using a set of very similar trajectories of the ball.

\section{Methodology}
\label{sec:format}
\begin{figure}[ht]
	\centering
	\includegraphics[width=\linewidth]{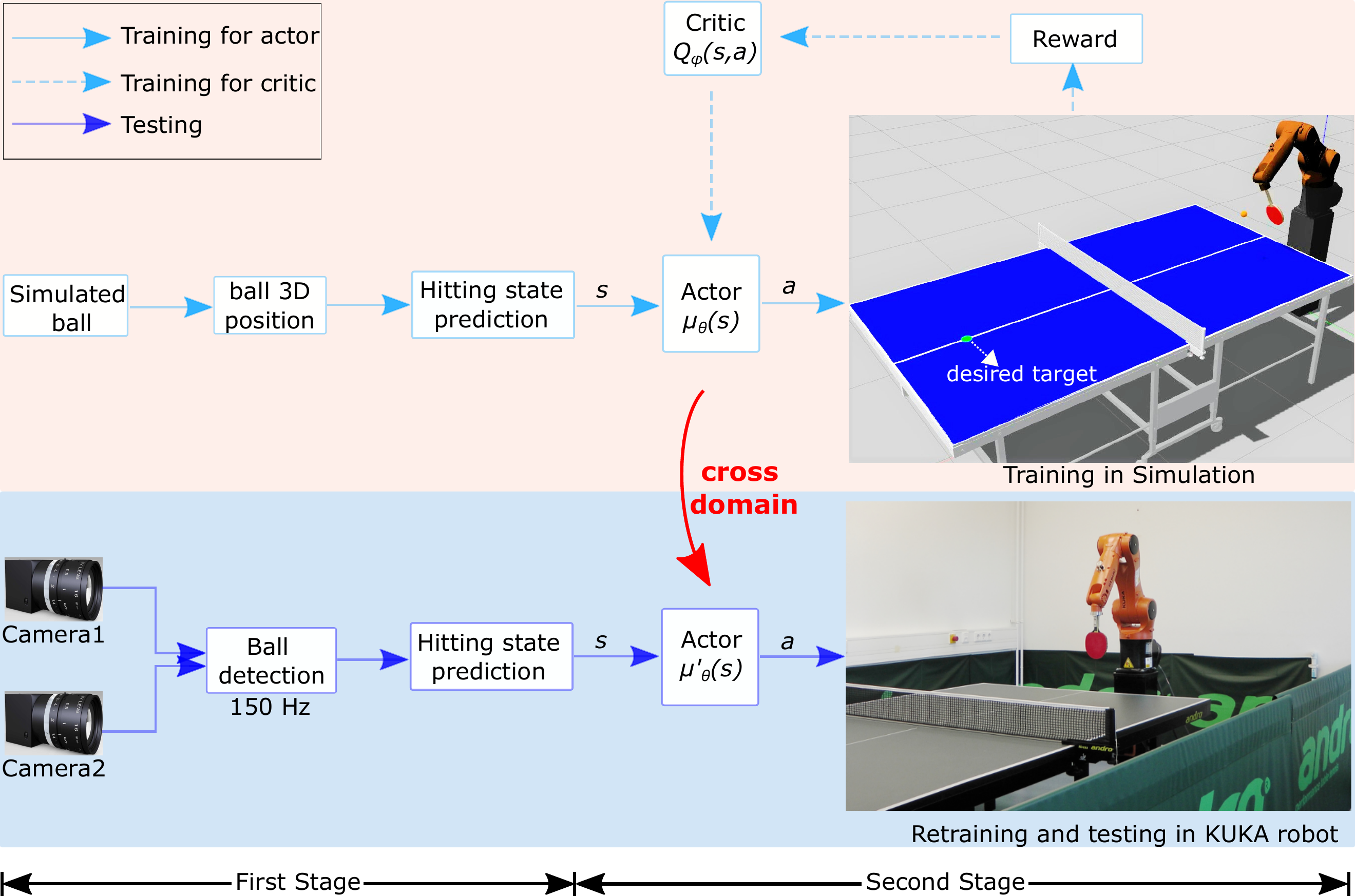}
	\caption{The entire framework for training and testing. The goal is returning the ball to the desired target position on the table. In the first stage, the ball's state $s$ at the hitting time is predicted by~\cite{tebbe2018table,tebbe2020spin}. In this work, we focus on the second stage, where an optimal stoke can be learned based on a novel RL algorithm. The upper part is performed in simulation. The RL model is trained within 10,000 episodes. To cross the reality gap, the model is then retrained in reality (see the bottom part).}
	\label{fig:framework}
\end{figure}
\begin{figure}[ht]
	\centering
	\includegraphics[width=0.85\linewidth]{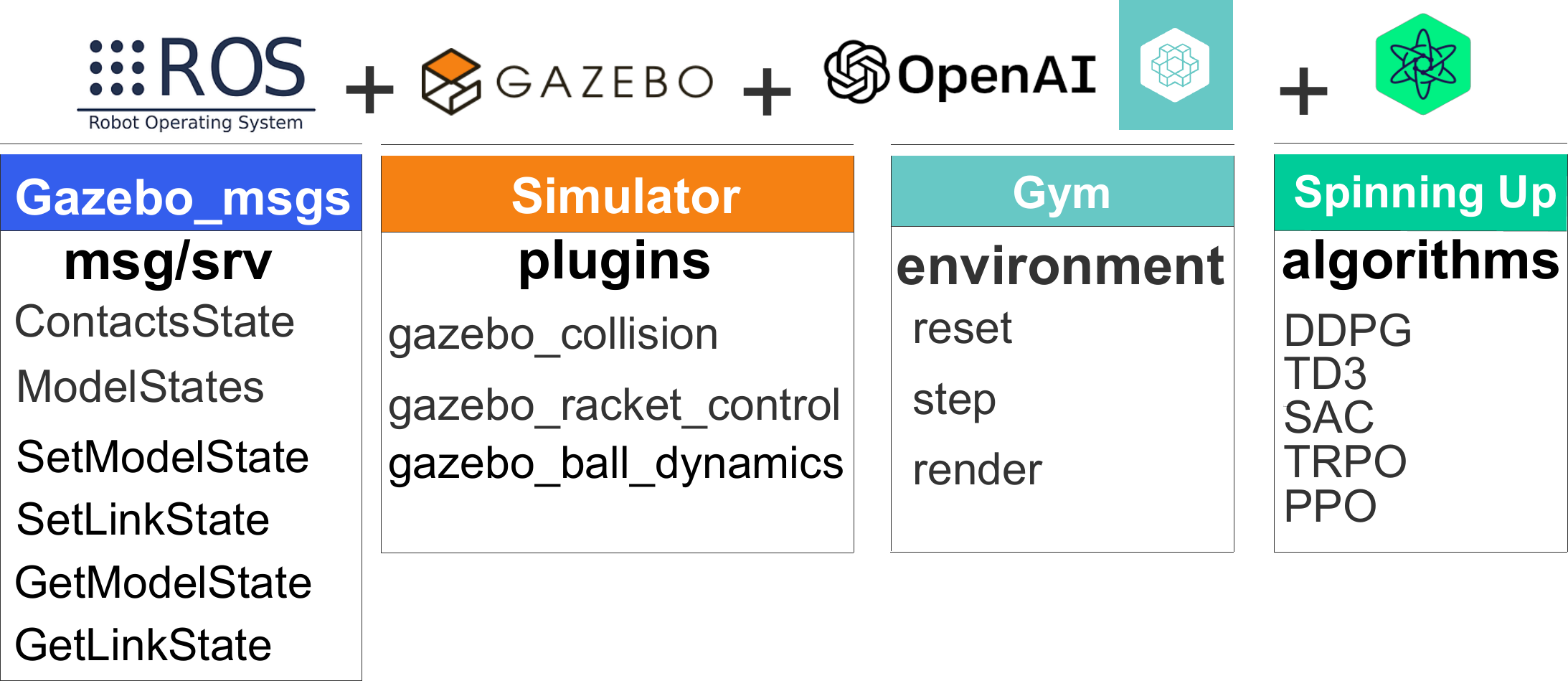}
	\caption{The learning architecture is built by combining the Gazebo simulator with ROS, Gym, and Spinning Up. We can subscribe and publish the states (pose and velocity) of the ball and the robot from Gazebo with the \texttt{Gazebo\_msgs} in ROS. These states should be fed into the Gym toolkit, which allows us to use the RL algorithms in Spinning Up.}
	\label{fig:arch}
\end{figure}

To efficiently learn the optimal stroke and successfully return the ball to the desired target position on the table, we propose a novel framework as shown in Fig.~\ref{fig:framework}. Specifically, a realistic simulation environment is developed for robot learning and comparison with other advanced RL algorithms. In the first stage, the hitting state (position, velocity, and spin) of the ball is predicted using the physical model of the ball proposed in our previous work~\cite{tebbe2018table, tebbe2020spin}. In the second stage, a novel approach is presented to learn the optimal stroke in simulation, which is conjugated with ROS and OpenAI libraries (see Fig.~\ref{fig:arch}).

\subsection{Simulation}

A well-known challenge for deep RL is the safe interaction with the environment. In particular, in robotic table tennis, it is difficult to explore all possibilities in realtiy, since unexpected collisions would destroy the mechanical robot parts. Moreover, the robot must interact with the environment over a large number of steps to learn a high-level policy. Taken together, this makes the application of deep RL in robotic table tennis more challenging.
To address these issues, we develop a realistic simulation that provides a convenient scenario for optimal stroke learning as well as a comparison of different algorithms. The pose and velocity of the simulated racket are controlled by the Gazebo plugins. The dynamic model for each entity is obtained with the methods described in the following subsections.

\subsubsection{Flying ball model}
In addition to the gravitational force $F_g$, a flying ball is usually influenced by the Magnus force $F_m$ and the air drag $F_d$~\cite{zhang2014}. As shown in Fig.~\ref{fig:force}, $F_m$ is perpendicular to the spin axis and the flight direction, while $F_d$ is opposite to the flight direction. These forces can be computed using the following formulas:

\begin{figure}[ht]
	\centering
	\includegraphics[width=0.55\linewidth]{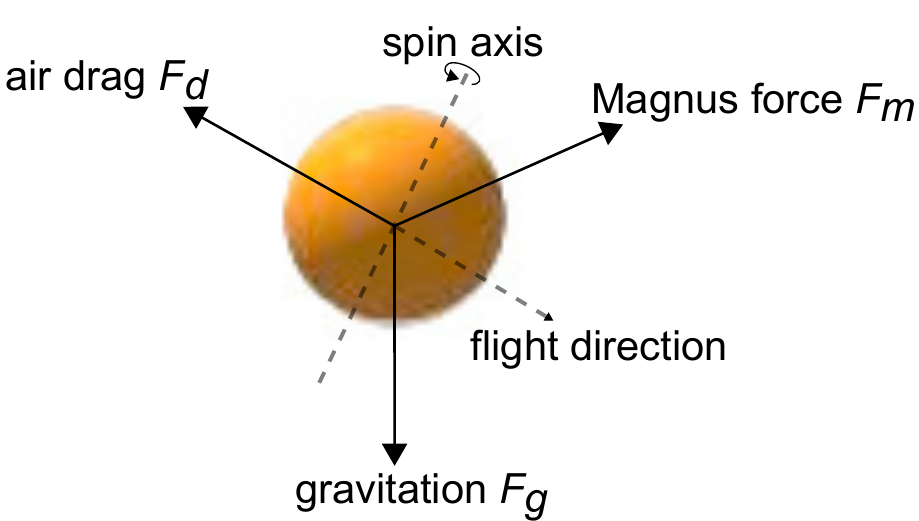}
	\caption{Force analysis of a flying ball. A sphere shell of radius $r_{1}$ and mass $m$, with a centered spherical cavity of radius $r_{2}$, is created as a simulated ball in Gazebo.}
	\label{fig:force}
\end{figure}

\begin{align}
F_g &= ( 0, 0, -mg)^T \,, \\ 
F_d &= - \frac{1}{2} C_D \rho_a A \norm{v} v \,, \\
F_m &= \frac{1}{2} C_M \rho_a A r_1 ( \omega \times v) \,,
\end{align}
where the constants were determined in our previous work~\cite{tebbe2018table}, including the mass of the ball $m=2.7 \text{g}$, the gravitational constant $g = 9.81 \text{m}/\text{s}^2$, the drag coefficient $C_D = 0.4$, the density of the air $\rho_a = 1.29 \text{kg}/\text{m}^3$, the lift coefficient $C_M = 0.6$, the radius of the ball $r_1 = 20 \text{mm}$, the cavity radius $r_2 = 19.6 \text{mm}$, and the cross-section of the ball $A = r_1^2 \pi$. In addition, $\omega$ and $v$ are the linear and angular velocities, respectively, which can be derived from the trajectory of the ball in reality. These two velocities can be further used to predict the hitting state of the ball using the algorithm introduced in ~\cite{tebbe2018table,tebbe2020spin}.
To simulate the accurate dynamics of the ball, the inertia value $I$ should also be considered, which is calculated as
\begin{align}
I&=\frac{2}{5} m\left(\frac{r_{1}^{5}-r_{2}^{5}}{r_{1}^{3}-r_{2}^{3}}\right)\,.
\end{align}

\subsubsection{Bounce model}
Since the physical contact between two objects in reality is a very complex matter, the Open Dynamics Engine (ODE), a popular rigid body dynamics library for robotics, was used to simulate the contact forces between the ball and the table (or the racket) in simulation.
ODE has been originally integrated into the Gazebo simulator. 
To represent the elastic and frictional impacts on the ball, we compute the restitution coefficient $\kappa_R$ and the friction coefficient $\mu$ similar to~\cite{blanksmart}.

The restitution coefficient $\kappa_R$ is defined as the ratio of the energy before and after a collision, e.g., when the ball bounces off the table. Approximately, it can be solved by the free fall of the ball as follows:
\begin{align}
    \kappa_{R}^{t}=\frac{v_{2}^{t}-v_{2}^{b}}{v_{1}^{b}-v_{1}^{t}}=\frac{-v_{2}^{b}}{v_{1}^{b}}=\frac{-\sqrt{2 \cdot g \cdot h_{2}}}{-\sqrt{2 \cdot g\cdot h_{1}}}=\sqrt{\frac{h_{2}}{h_{1}}} \,,
\end{align}
where $v_{1}^{b}$ and $v_{1}^{t}$ are the velocities of the ball and table before bouncing, while $v_{2}^{b}$ and $v_{2}^{t}$ are the velocities after bouncing. $h_1$ and $h_2$ are the corresponding heights when the ball is not moving. Here the table velocity $v^t=0$.

\begin{figure}[ht]
	\centering
	\includegraphics[width=0.3\linewidth]{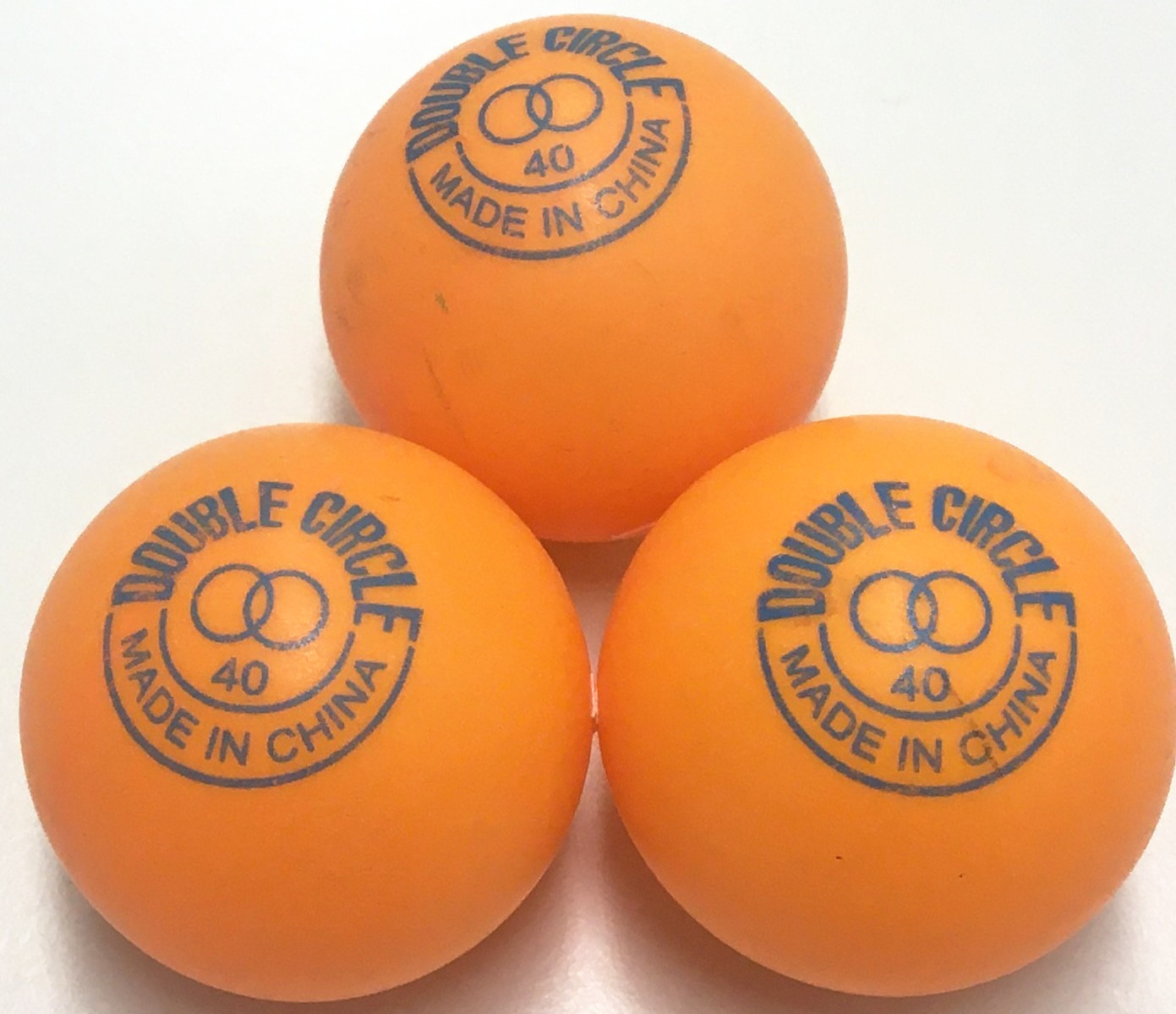}
	\caption{Setup for measuring the friction coefficient $\mu$. Three balls are arranged together in the shape of a triangle frame and placed on a flat table.}
	\label{fig:friction}
\end{figure}

The friction coefficient $\mu$ is obtained from the setup in Fig.~\ref{fig:friction}, where three balls are arranged together in the shape of a triangle frame. We first place the balls on the table and then lift the table until the balls start to slide. After obtaining the horizontal angle change $\theta$ of the table, the friction coefficient $\mu^t$ between the table and the ball can be calculated as 
\begin{align}
    \mu^t=\frac{3mg \cdot \sin{\theta}}{3mg \cdot \cos{\theta}}=\tan{\theta} \,.
\end{align}

With the same methods, we calculate the restitution coefficient $\kappa_R^r$ and the friction coefficient $\mu^r$ of the racket. The resulting parameter values are given in the Table~\ref{table:contact}. 
The additionally required parameters $\mu_2$ and $slip$ are defined as the friction coefficient in the second ODE friction pyramid direction and the coefficients of force-dependent-slip (FDS), respectively. They are manually adjusted to fit the reality.

\begin{table}[h]
	\caption{Collision parameters when the ball impacts on the table and racket.}
	\begin{center}
		\begin{tabular}{l|cccc}
			\hline
			& $\kappa_R$ & $\mu$  & $\mu_2$  & $slip$ \\
			\hline\hline
			Table & 0.97 & 0.05 & 0.025 & 0.01\\
			\hline
			Racket & 0.9  & 1.0 & 0.025 & 0.01 \\
			\hline
		\end{tabular}
	\end{center}
	\label{table:contact}
\end{table}

\begin{figure}[ht]
	\centering
	\includegraphics[width=0.9\linewidth]{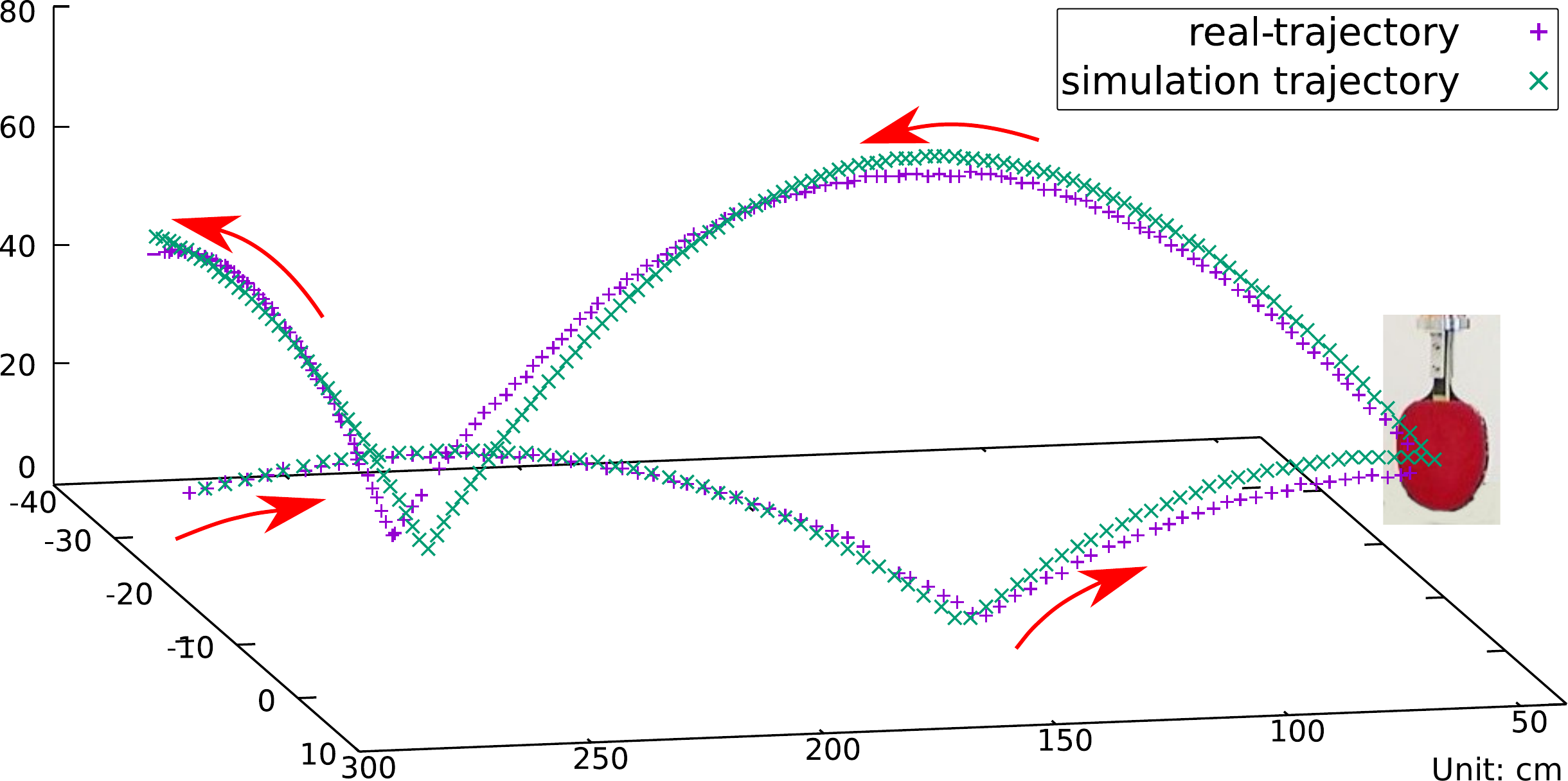}
	\caption{Trajectory comparison between reality and simulation when serving a topspin ball. The difference between the returned landing positions on the table is about 6.2 cm. In this case, the racket is stationary since it is difficult to understand and simulate all the dynamics parameters of a moving racket in simulation. Therefore, the actions of the racket applied in simulation are always the true actions without any noise and time latency. This reality gap is closed in the retraining section~\ref{subsec:retraining}.}
	\label{fig:topspin}
\end{figure}

To roughly test the accuracy of the simulation, we utilize a ball throwing machine to launch a topspin ball towards the stationary racket mounted on the robot. The entire trajectory of the ball can be recorded as the ground truth using stereo cameras at 150 Hz. The starting spin and velocity of the ball are computed using a spin detector tool and a curve fitting approach, respectively.  Then, these initial parameters are fed into the simulated ball and this initialized ball is served in simulation. The simulated trajectory is generated and is shown in Fig.~\ref{fig:topspin}. 

\subsection{Algorithm}
With regard to the different types of inputs, there are generally two ways available when using deep RL algorithms in robotic table tennis. 
One is the one-stage algorithm that takes the state of the ball at each step as input and learns the pose of the racket in an end-to-end way. The other is the two-stage algorithm that first predicts the hitting state (i.e. position, velocity, and spin) of the ball and then takes it as input for RL. The latter can significantly accelerate the training step and can efficiently deal with different spin balls. In this work, we adopt the second approach to learn the optimal stroke of the racket based on the prediction of the state of the ball at the hitting time.

Since there is only a single state vector as input in the second stage, we then parameterize stroke learning as a bandit problem, where actions have no influence on the next states and consequently there are no delayed rewards in an episode. It is a simple version of an Markov Decision Process (MDP), with 
\begin{align}
    M=(S, A, R) \,,
\end{align}
where $S$ is the set of the observed 11-D states $s$, including the 3D position $p^b$ of the ball, 3D linear velocity $v^b$, 3D angular velocity $\omega^b$ at the hitting time, and the desired 2D landing target position $p^{tar}$ on the table. $A$ is the set of 3D actions $a$ that can be performed on the robot. Due to the restriction of the current mechanical structure and the control system, we cannot operate the robot as flexibly as a human can move. Therefore, we only learn to change the linear velocity $v_x^r$ of the robot along the $x$-axis and its orientation angles ($\beta^r, \gamma^r$) around the $y$ and $z$ axes. The target position of the racket is the same as the predicted hitting position of the ball.
$R$ is a set of the immediate rewards $r$.

\begin{figure}[ht]
	\centering
	\includegraphics[width=0.9\linewidth]{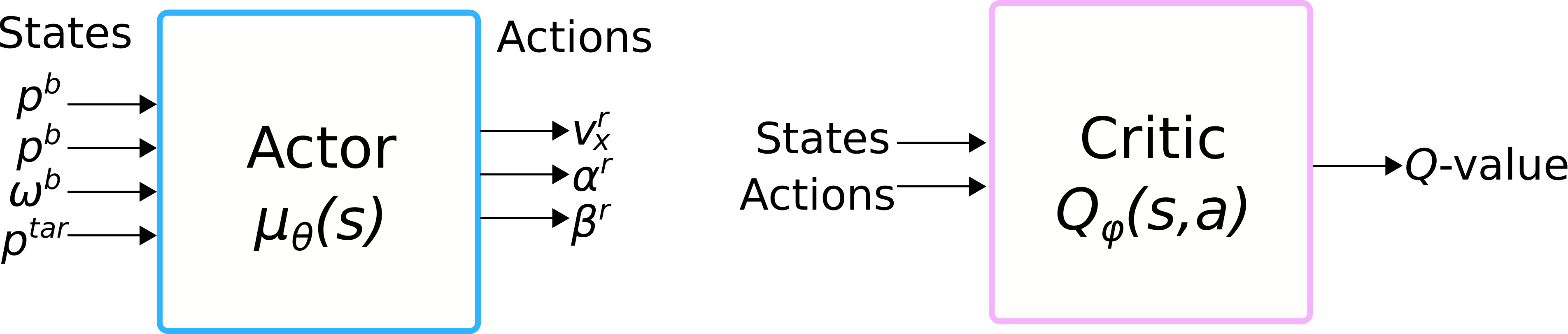}
	\caption{Classic Actor-Critic algorithms. Instead of a 1D Q-value, a 3D Q-value is proposed to train the corresponding 3D actions.}
	\label{fig:ac}
\end{figure}

We use a policy $\mu_\theta(s)$ as the actor network which can output the actions $a$ with respect to the current state $s$, as shown in Fig.~\ref{fig:ac} left. To evaluate the actions, a critic network $Q_{\phi}(s, a)$ is used, which takes as input both the states and actions and outputs a $Q$-value vector, as shown in Fig.~\ref{fig:ac} right. $\theta$ and $\phi$ are the weights of the neural network. The goal of our work is to lean a deterministic policy $\mu_\theta(s)$, which provides an action that maximizes the norm of $Q_{\phi}(s, a)$. According to the DDPG algorithm, the critic and the actor can be updated, respectively, by minimizing the losses:

\begin{align}
\quad \quad \quad \mathcal{L}(\phi, \mathcal{B})&=\underset{\left.(s, a, r\right.) \sim \mathcal{B}}{\mathrm{E}}\left.\norm{Q_{\phi}(s, a)-r}^{2}\right. \,,
\label{eq:critic_loss}
\end{align}

\begin{align}
    \mathcal{L}(\theta, \mathcal{B})=\underset{s \sim B}{-\mathrm{E}}\left.\norm{Q_{\phi}(s, \mu_{\theta}(s)}\right. \,,
    \label{eq:actor_loss}
\end{align}
where $B$ is a minibatch for storing $s,a,r$. The reward  $r$ is the feedback from the environment, which we will discuss in more detail later.

To accelerate the training step and boost the resulting performance, we apply the following modifications to the classic actor-critic algorithms for training.

\subsubsection{Exploration}
For continuous action spaces, several exploration strategies are used in the deterministic environments. For instance, the $random$ strategy selects actions randomly from a Gaussian distribution, while the $epsilon-greedy$ strategy takes the random actions occasionally with probability $\epsilon$ and uses the output of the current actor $\mu_\theta(s)$ with probability $1-\epsilon$. This is the default strategy used in Spinning Up.
In this work, we noticed that the actor $\mu_\theta(s)$ did not give the action with the maximum $Q$-value in the earlier training step because of the large loss error. Therefore, we generate the action $a$ as 

\begin{align}
    a=\operatorname*{argmax}_{\mu_\theta(s)+\mathcal{N}} \norm{Q_\phi(s, \mu_\theta(s)+\mathcal{N})} \,,
\label{eq:noise}
\end{align}
where $\mathcal{N}$ is a Gaussian noise.

\subsubsection{Reward shaping}
In~\cite{zhu2018}, a reward function was developed that depends on the height $h^b$ of the ball when crossing the net and the actual landing position $p^{real}$ on the table when the ball is returned. 
To balance these two independent variables $h^b$ and $p^{real}$, a coefficient is used; however, deciding the value of this coefficient is a tricky issue.
To address this problem, we decompose the reward function into three vector components: $r_x$ and $r_y$ for the $x$ and $y$ of the landing position, and $r_h$ for the height of the ball. Each reward function is then normalized to [0,1] by the following equations: 
\begin{align}
    r_x &= e^{-\abs{p_x^{real}-p_x^{tar}}} \,, \\
    r_y &= e^{-\abs{p_y^{real}-p_y^{tar}}} \,, \\ 
    r_h &= e^{-\abs{h^b-0.173}} \,, \\
    r &= [r_x, r_y, r_h] \quad \text{if \textit{success} else} \overrightarrow{\text{0}} \,,
\end{align}
where $p_x$ and $p_y$ are the landing position (in meters) along the $x$ and $y$ axes, the constant 0.173 is the measured actual height of the net (in meters), and $e$ is the natural exponential operation. When a ball is successfully returned to the opposite table, $success$ is set to true. 

\subsubsection{3D $Q$-value}
Normally, the $Q$-value is a 1D vector that is expected to be maximized. To take advantage of the above rewards, we replace the last layer in the critic network from 1D to 3D. This results in a 3D $Q$-value $[Q_x, Q_y, Q_h]$, which can precisely indicate the quality of the actions. 

In addition, for the actor-critic model we adopt TD3 as the backbone. The critic is changed to:
\begin{align}
Q_{\phi}(s, a)=\left\{\begin{array}{l}
Q_{\phi_1}(s, a), \text { if }\norm{Q_{\phi_1}(s, a)} < \norm{Q_{\phi_2}(s, a)} \\
Q_{\phi_2}(s, a), \text { otherwise }\\
\end{array}\right. \,.
\label{eq:15}
\end{align}

The whole training process is depicted in Algorithm~\ref{algorithm:1}, where the loss functions $\mathcal{L}(\phi_i, \mathcal{B})$ and $\mathcal{L}(\theta, \mathcal{B})$ are used to update the critic and the actor, respectively.
 \begin{algorithm}
 \caption{Policy Gradient Training with TD3 backbone}
 \begin{algorithmic}[1]
 \renewcommand{\algorithmicrequire}{\textbf{Input:}}
 \renewcommand{\algorithmicensure}{\textbf{Output:}}
 \Require Initial actor weights $\theta$, critic weights $\phi_1$ and $\phi_2$, replay buffer $\mathcal{D}$, number of episodes $\lambda$, Gaussian noise $\mathcal{N}(0,0.1)$
 \Ensure  Optimal policy $\mu^*_\theta(s)$
 \For{ n=1 to $\lambda$ }
 \State Observe the state $s$ and generate the action $a$ by \newline
 \State  \quad \quad \quad    $\operatorname*{argmax}_{\mu_\theta(s)+\mathcal{N}} \norm{Q_\phi(s, \mu_\theta(s)+\mathcal{N}})$ \newline
 \State Apply $a$ in the environment and get the reward $r$
 \State Store ($s,a,r$) in the replay buffer $\mathcal{D}$
 \State Reset the environment
 \If{it is time to update}
 \State Sample a random minibatch $\mathcal{B}$ from $\mathcal{D}$
 \For{i=1,2}
 \State Update the critic by minimizing the loss: \newline
 \State \quad  $\mathcal{L}(\phi_i, \mathcal{B})\&=\underset{\left(s, a, r\right) \sim \mathcal{B}}{\mathrm{E}}\left[\norm{Q_{\phi_i}(s, a)-r}^{2}\right]$ \newline
 \EndFor
  \If {it is time to update actor}
  \State Update the actor by  minimizing the loss: \newline
  \State \quad $\mathcal{L}(\theta, \mathcal{B})\&=\underset{s \sim B}{-\mathrm{E}}\norm{Q_{\phi}\left(s, \mu_{\theta}(s)\right)}$ \,, \newline
  \State where $Q_{\phi}(s, \mu_{\theta}(s))$ is got from Eq.~\ref{eq:15}:
  \EndIf
 \EndIf
 \EndFor
 \end{algorithmic}
 \label{algorithm:1}
 \end{algorithm}

\section{Experiments}
\label{sec:experiments}
\subsection{Training and testing}

To generalize the trained model, 11,000 serves were randomly sampled from a wide range of values for training (10,000) and evaluation (1,000), as shown in Table~\ref{table:dataset}.
To bridge the reality gap, a Gaussian noise is first applied to the 3D position of each ball in simulation. Then, we predict the state of the ball at the hitting time using the methods in~\cite{tebbe2018table, tebbe2020spin} instead of using the actual state of the ball in simulation. Since the predicted hitting position is the one to which the simulated racket should actually move, this allows the replication of the real situation and makes the trained model more realistic for the real world.
The final state variables and their range of the ball are shown in Table~\ref{table:dataset}, which includes the desired landing target position ($p_x^{tar},p_y^{tar}$), the position of the ball ($p_x^b,p_y^b,p_z^b$), linear velocity ($v_x^b, v_y^b, v_z^b$) and angular velocity ($\omega_x^b, \omega_y^b, \omega_z^b$) at the hitting time. These state variables are then normalized and used as inputs for training. The $p_x^b$ is consistently equal to 0.675 m to form a virtual hitting plane used for the state prediction in the first stage. Thus, the hitting time is the time when the ball reaches this virtual plane. The home position of the racket in the world coordinates is $(0.5, 0.0, 0.0)$ in meters.

\begin{table}[h]
	\caption{State variables and their range at the hitting time for training and evaluation in simulation.}
	\begin{center}
		\begin{tabular}{c|cc}
			\hline
			& training&evaluation \\
			\hline\hline
			$p_x^{tar}$ & \multicolumn{2}{c}{2.55m} \\
			\hline
			$p_y^{tar}$ & \multicolumn{2}{c}{0.0m} \\
			\hline
			$p_x^b$ & \multicolumn{2}{c}{0.675m} \\
			\hline
			$p_y^b$ & \multicolumn{1}{c}{[-0.60m, 0.63m]} & [-0.68m, 0.68m] \\
			\hline
			$p_z^b$ & \multicolumn{1}{c}{[-0.01m, 0.34m]} & [-0.01m, 0.34m] \\
			\hline
			$v_x^b$ & \multicolumn{1}{c}{[-6.00m/s, -1.35m/s]} & [-5.94m/s, -2.52m/s] \\
			\hline
			$v_y^b$ & \multicolumn{1}{c}{[-1.95m/s, 2.16m/s]} & [-1.29m/s, 2.02m/s] \\
			\hline
			$v_z^b$ & \multicolumn{1}{c}{[-3.47m/s, 3.15m/s]} & [-3.40m/s, 2.60m/s] \\
			\hline
			$\omega_x^b$ & \multicolumn{1}{c}{[-127.67rad/s, 110.88rad/s]} & [-95.08rad/s, 111.53rad/s] \\
			\hline
			$\omega_y^b$ & \multicolumn{1}{c}{[-299.99rad/s, 299.81rad/s]} & [-299.62rad/s, 299.73rad/s] \\
			\hline
			$\omega_z^b$ & \multicolumn{1}{c}{[-193.81rad/s, 189.65rad/s]} & [-189.05rad/s, 189.47rad/s] \\
			\hline
			\hline
			Episodes $\lambda$ & \multicolumn{1}{c}{10000} & 1000\\
			\hline
		\end{tabular}
	\end{center}
	\label{table:dataset}
\end{table}

Considering the mechanical setup of the robot, we restrict the linear velocity $v_x^r$ of the robot to a range from 0m/s to 2m/s. The orientation angles $\beta^r, \gamma^r$ are between \ang{-50} to \ang{50}. The third angle $\alpha^r$ around the $x$-axis is calculated as
\begin{align}
    \alpha^r = k\cdot \frac{p^b_y}{0.5\cdot w^t} \,,
\end{align}
where $w^t$ is the table width and $k$ is a weight. In this way, the robot generates a flexible stroke. Here we assume the angle $\alpha^r$ has no influence on the impact with the ball. 

In Equation~\ref{eq:noise}, the added action noise $\mathcal{N}$ for exploration is a mean-zero Gaussian distribution with a standard deviation of 0.1. The replay buffer $\mathcal{D}$ has a size of 5,000. The number of training episodes $\lambda$ is 10,000. Other hyper-parameters used for actor-critic are given in Table~\ref{table:hyper}. The output actions from the actor are scaled to the valid range and then applied to the simulation. These hyper-parameters are tuned manually in order to achieve the best performance.
\begin{table}[h]
	\caption{Hyper-parameters used for training in simulation and retraining in reality.}
	\begin{center}
		\begin{tabular}{l|c|c}
			\hline
			& \multicolumn{2}{c}{Actor/Critic} \\
			\hline
			& Training & Retraining\\
			\hline\hline
			batch size & 512 & 50\\
			\hline
			epochs  & 100 & -\\
			\hline
			episodes per epoch  & 100 & 20\\
			\hline
			learning rate  & 1e-4 & 5e-5 \\
			\hline
			optimizer & \multicolumn{2}{c}{Adam} \\
			\hline
			layers & \multicolumn{2}{c}{[256,256,3]} \\
			\hline
			activation  & \multicolumn{2}{c}{relu} \\
			\hline
			output activation  & \multicolumn{2}{c}{tanh/linear} \\
			\hline
		\end{tabular}
	\end{center}
	\label{table:hyper}
\end{table}

\begin{figure}[tb!]
\begin{center}
\begin{tabular}{c}
\includegraphics[width=0.85\textwidth]{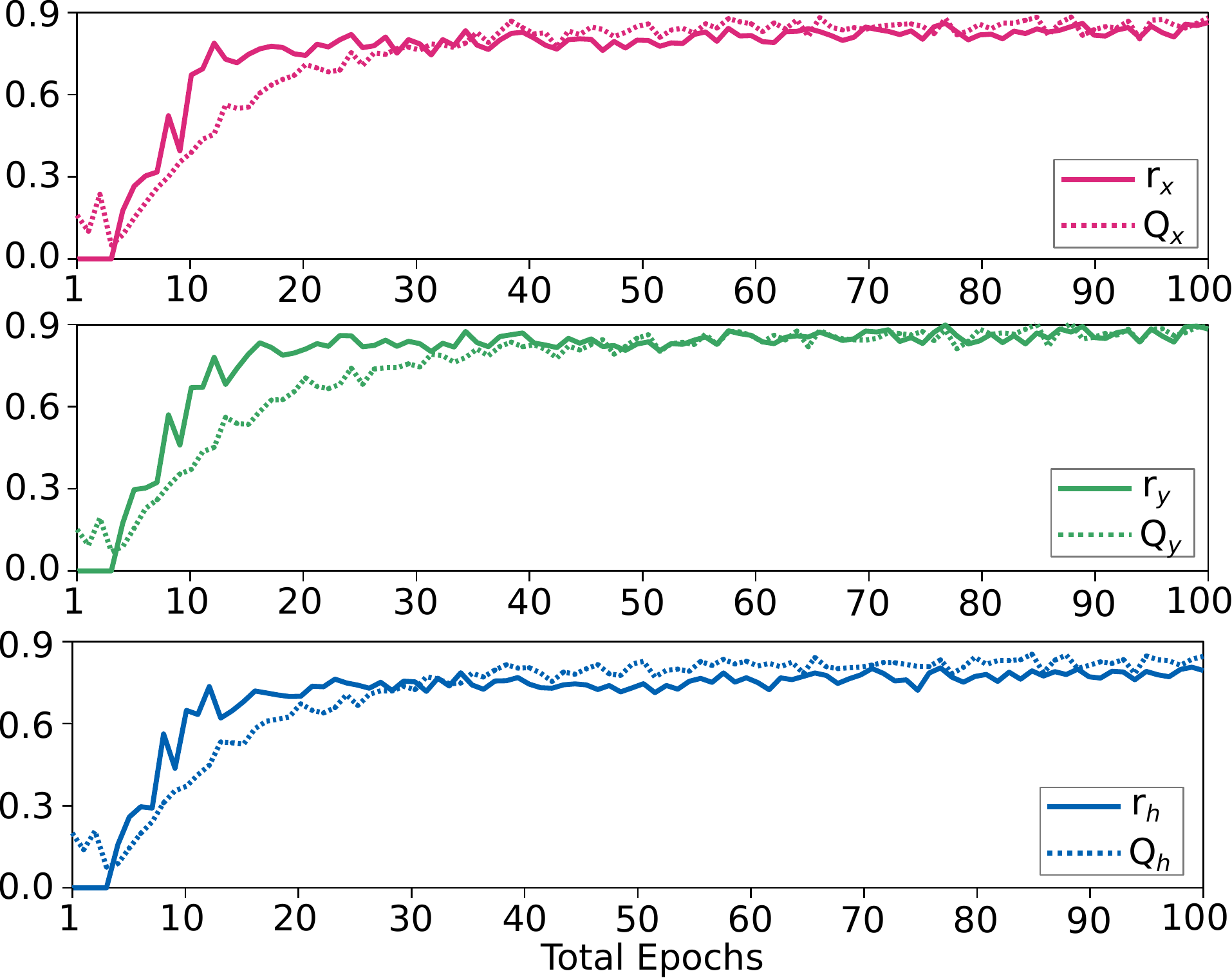}
\end{tabular}
\end{center}
   \caption{The testing rewards $[r_x, r_y, r_h]$ and the 3D $Q$-value $[Q_x, Q_y, Q_h]$ over the epochs in simulation.}
\label{fig:test_rews} 
\end{figure}

Compared to other environments that require millions of interactions in OpenAI Gym, our model is able to converge after 30 epochs, which took about 1 hour of training.  
In addition, 1000 episodes were run for evaluation after each epoch. The resulting rewards and the corresponding 3D $Q$-value are plotted in Fig.~\ref{fig:test_rews}. It is observed that the testing rewards reach a stable level starting from the $20^{th}$ epoch, although the $Q$-values have not yet converged to the maximum values.  

\subsection{Evaluation}
A commonly used metric for evaluation is the distance error between the actual and desired landing position~\cite{buchler2020learning,mahjourian2018hierarchical,zhu2018}. However, this metric cannot reflect a failed return, e.g., if the landing position is not on the table, then it will be difficult to calculate the distance error. Therefore, we introduced a new metric here: distance error $\epsilon_d$ computed by
\begin{align}
    r_d &=  e^{-\norm{p^{real}-p^{tar}}_2} \quad \text{if \textit{success} else 0} \,, \\ 
    \epsilon_d &= -\ln \left( \frac{1}{\lambda}\sum_{n=1}^{\lambda} r_d^n \right) \,,
\end{align}
where the $r_d^n$ is the reward for the landing distance error of the $n^{th}$ ball. The value of the $r_d^n$ is 0 if the ball does not land on the opposite table.
Furthermore, two additional metrics reflecting the performance of the model were used for evaluation. The flying height error $\epsilon_h$ of the ball when crossing the net is calculated by
\begin{align}
    \epsilon_h &= -\ln \left( \frac{1}{\lambda}\sum_{n=1}^{\lambda} r_h^n \right) \,.
\end{align}
As well, the success rate of the ball in returning to the opposite table is calculated.
To obtain a fair evaluation, we adopt 1000 episodes to cover a large range for the states (see Table~\ref{table:dataset} right).

Since only the 1D $Q$-value was used in the existing RL algorithms, we thus create a 1D reward function similar to~\cite{zhu2018} as
\begin{align}
    r_{eval} = e^{-k(\norm{p^{real}-p^{tar}}-\abs{ h^b-0.173})} \,,
\end{align}
where $k$ is a scalar coefficient set to 0.5 in this work. This new reward function is used only for training the existing RL algorithms, including TRPO, PPO, SAC, DDPG, and TD3. 
By evaluating the existing and newly introduced algorithms using different reward functions corresponding to these different algorithms, 
we compute the distance error $\epsilon_D$, the height error $\epsilon_h$, and the success rate, respectively, as shown in Table~\ref{table:eval}. The unit of these errors is converted from meters to centimeters for better visualization. The proposed approach, argmax exploration plus 3D $Q$-value together with TD3 backbone, achieves better performance than the DDPG backbone. The other three approaches, TRPO, PPO, and SAC, learn the optical stroke using a stochastic policy, resulting in much higher errors and lower success rate.
\begin{table}[h]
	\caption{Evaluation for different Algorithms.}
	\begin{center}
		\begin{tabular}{ >{\arraybackslash}p{4cm}|ccc}
			\hline
			Algorithms & $\epsilon_d$  & $\epsilon_h$ & success rate  \\
			\hline\hline
			TRPO  &  47.0cm&31.0cm&84.8\%\\
			\hline
			PPO  &  44.2cm&30.8cm&87.1\%\\
			\hline
			SAC &  43.5cm&29.0cm&89.2\%\\
			\hline
			DDPG &  25.6cm& 22.1cm&95.6\%\\
			DDPG+argmax &  23.0cm& 22.3cm&97.4\%\\
			DDPG+argmax+3D Q-value &  21.3cm& 21.7cm&97.9\%\\
			\hline
			TD3 & 25.2cm& 22.3cm& 97.2\%\\
			TD3+argmax & 22.2cm & 21.2cm & 97.7\% \\
			TD3+argmax+3D Q-value  & \textbf{20.3cm} & \textbf{21.2cm} & \textbf{98.5\%}\\
			\hline
		\end{tabular}
	\end{center}
	\label{table:eval}
\end{table}

\subsection{Retraining in reality}
\label{subsec:retraining}

Although we built a high-fidelity simulation by manually measuring the coefficients and applying random noise to the ball, the real robot has many more dynamic and complicated factors that cannot be accurately measured and accounted for. 
To find the best hyper-parameters for retraining, we change the racket's restitution coefficient $\kappa_R^r$ and friction coefficient $\mu^r$ in simulation. In this way, we can replicate the situation between two different rackets in reality. Based on the pretrained actor-critic model, we then retrain the model in the new simulation with different batch sizes, episodes per epoch, and learning rates. The best hyper-parameters found in simulation are shown in Table~\ref{table:hyper} right.

\begin{figure}[ht]
\centering
\begin{tabular}{c}
\includegraphics[width=0.8\textwidth]{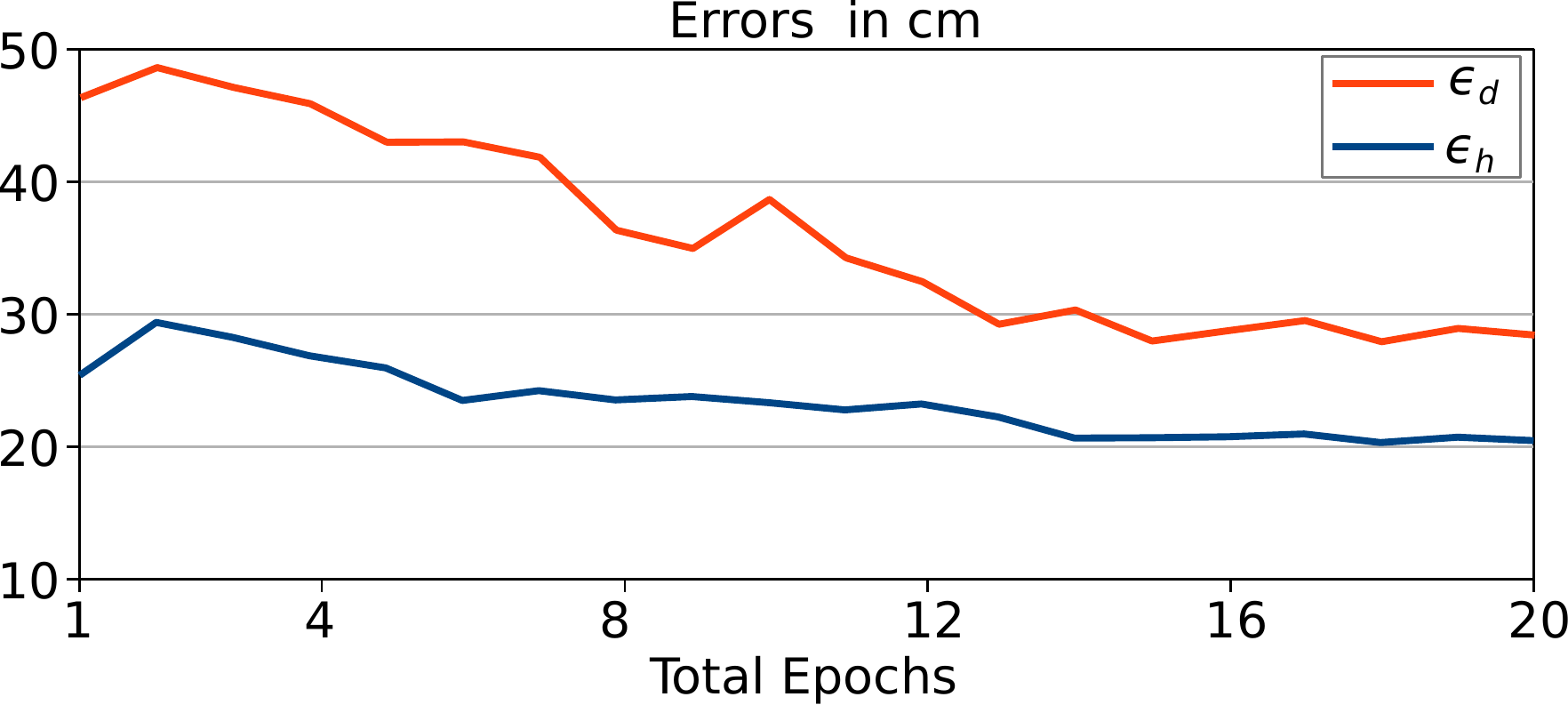} \\
(a) The first racket \\
\includegraphics[width=0.8\textwidth]{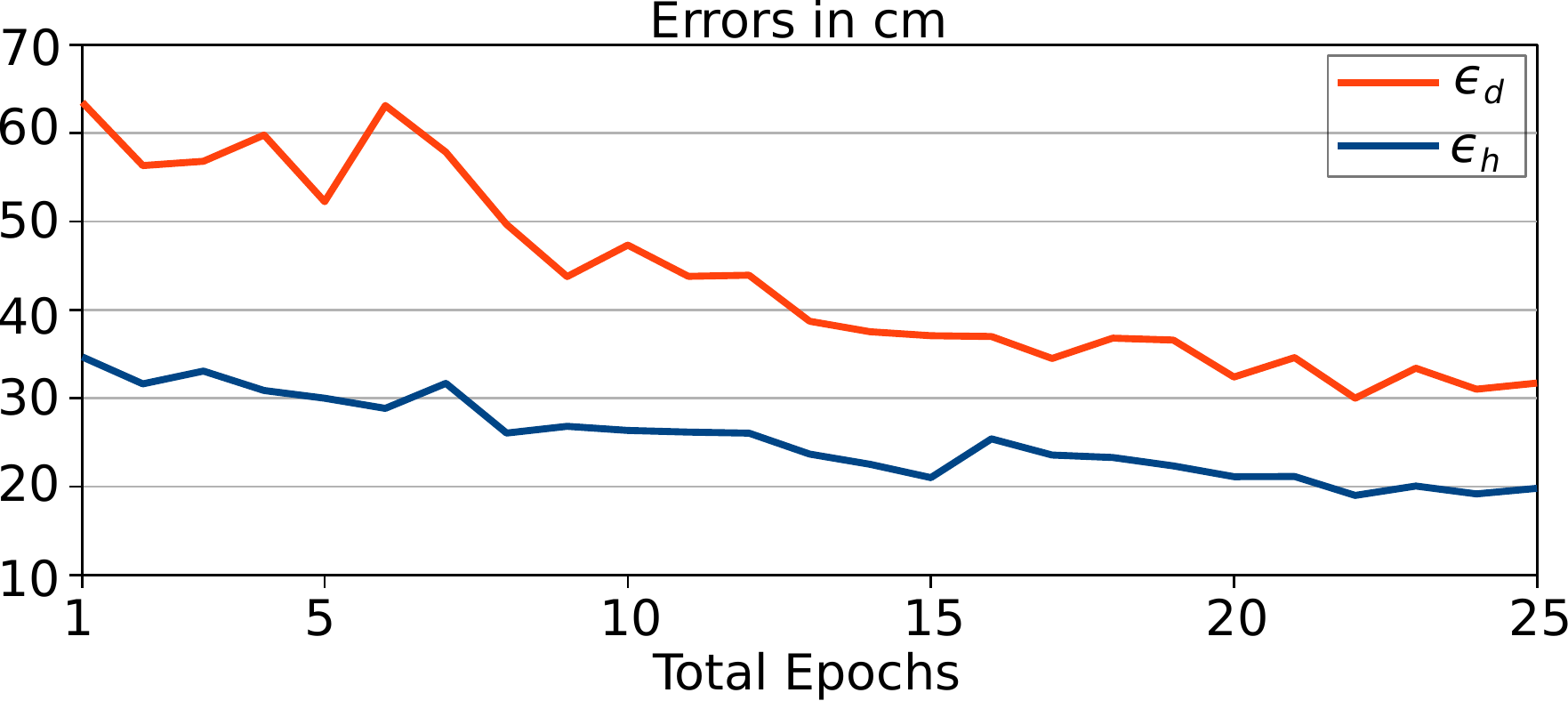}\\
(b) The second racket
\end{tabular}
\caption{Retraining process for the original racket (a) and for the coefficient-unknown racket (b), with balls served by a ball throwing machine. The metrics $\epsilon_d$ and $\epsilon_h$ in units of centimeters (cm) are the landing distance error $\epsilon_d$ and the height error $\epsilon_h$ when the ball crosses the net.}
\label{fig:rackets}
\end{figure}

A ball throwing machine, TTmatic 404A, is used to provide a variety of balls with sidespin, topspin, and backspin using a group of selected parameters.
At the moment our robot can only handle sidespin and topspin, since the backspin ball causes too much acceleration in a robot joint. This could be solved in the future. In addition, the Reflexxes motion library~\cite{kroger2010line} is used for robot trajectory planning in Cartesian space. Each epoch includes both sidespin and topspin balls during retraining. The state variables and the range of the ball for retraining and testing at the hitting time are shown in Table~\ref{table:range_kuka}. Here, the model is retrained with 20 epochs in 0.5 hours to ensure that it achieves convergence. The hitting position $p_x^b$ along the $x$ axis is fixed to 0.675 m. 
The retraining process demonstrating the landing distance error $\epsilon_d$ and the height error $\epsilon_h$ is shown in Fig.~\ref{fig:rackets}. 

\begin{table}[hbt!]
	\caption{State variables and their range at the hitting time for retraining and testing in reality.}
	\begin{center}
		\begin{tabular}{c|cc}
			\hline
			& retraining/testing in machine & testing with human\\
			\hline\hline
			$p_y^b$ & \multicolumn{1}{c}{[-0.55m, 0.64m]} & [-0.65m, 0.43m] \\
			\hline
			$p_z^b$ & \multicolumn{1}{c}{[0.085m, 0.34m]} & [0.06m, 0.0.33m] \\
			\hline
			$v_x^b$ & \multicolumn{1}{c}{[-5.20m/s, -3.5m/s]} & [-5.6m/s, -2.9m/s] \\
			\hline
			$v_y^b$ & \multicolumn{1}{c}{[-1.05m/s, 2.35m/s]} & [-2.38m/s, 1.25m/s] \\
			\hline
			$v_z^b$ & \multicolumn{1}{c}{[-0.78m/s, 3.92m/s]} & [-0.4m/s, 2.48m/s] \\
			\hline
			$\omega_x^b$ & \multicolumn{1}{c}{[-32.94rad/s, 52.68rad/s]} & [-33.00rad/s, 78.48rad/s] \\
			\hline
			$\omega_y^b$ & \multicolumn{1}{c}{[-210.52rad/s, 5.33rad/s]} & [-182.72rad/s, -55.28rad/s] \\
			\hline
			$\omega_z^b$ & \multicolumn{1}{c}{[-157.65rad/s, 34.51rad/s]} & [-66.68rad/s, 52.62rad/s] \\
			\hline
		\end{tabular}
	\end{center}
	\label{table:range_kuka}
\end{table}

Furthermore, to investigate the generalizability of the algorithms for a coefficient-unknown racket, we retrain a new model for a second racket whose dynamics are completely different from the first. 
Fig.~\ref{fig:rackets} (b) illustrates the second retraining process. As can be seen, the retraining of the second racket required around 5 more epochs to converge and achieved similar performance to the retraining of the first racket.

\subsection{Testing in reality}

\begin{figure}[ht]
	\centering
	\includegraphics[width=0.32\linewidth]{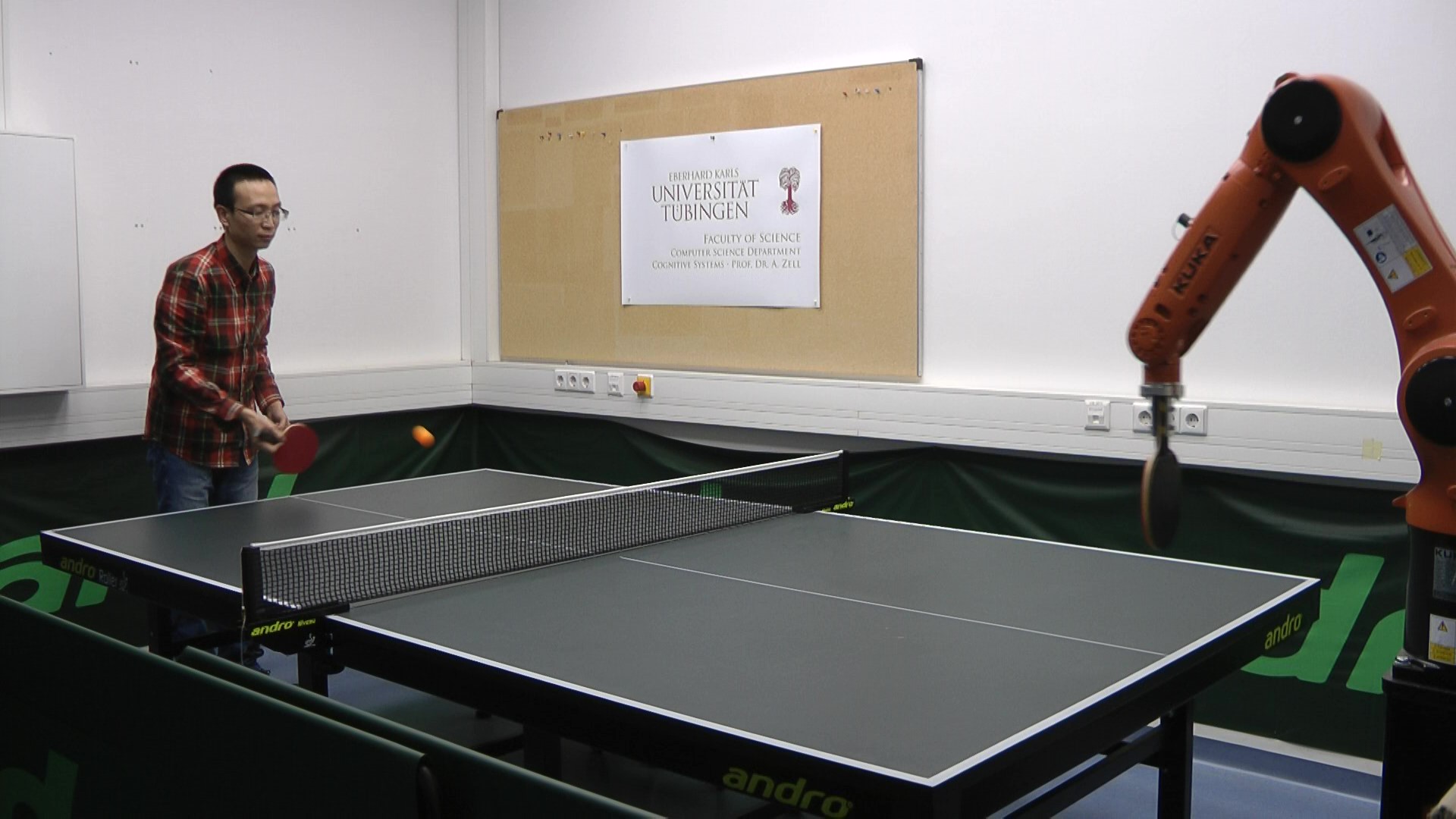}
	\includegraphics[width=0.32\linewidth]{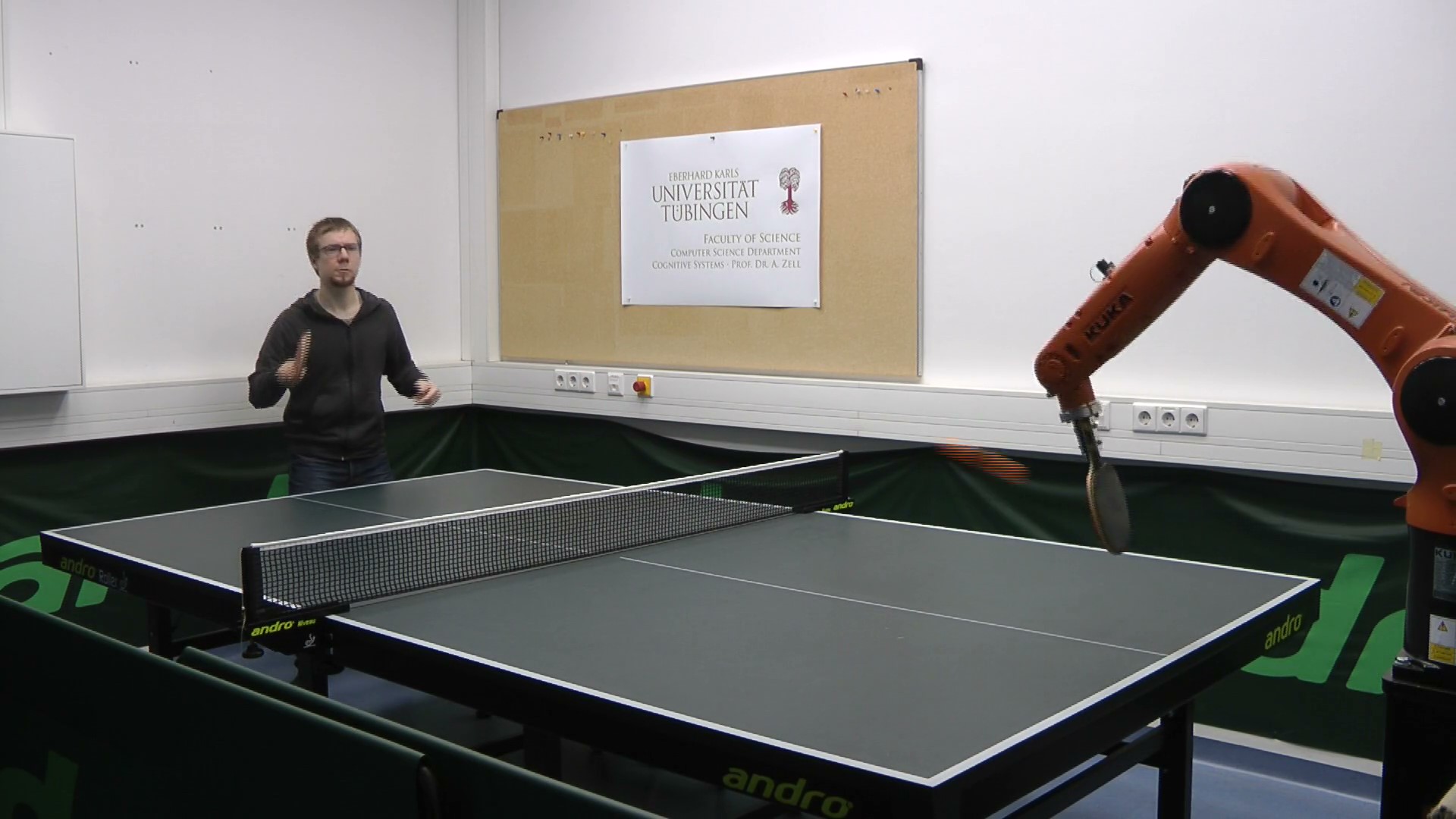}
	\includegraphics[width=0.32\linewidth]{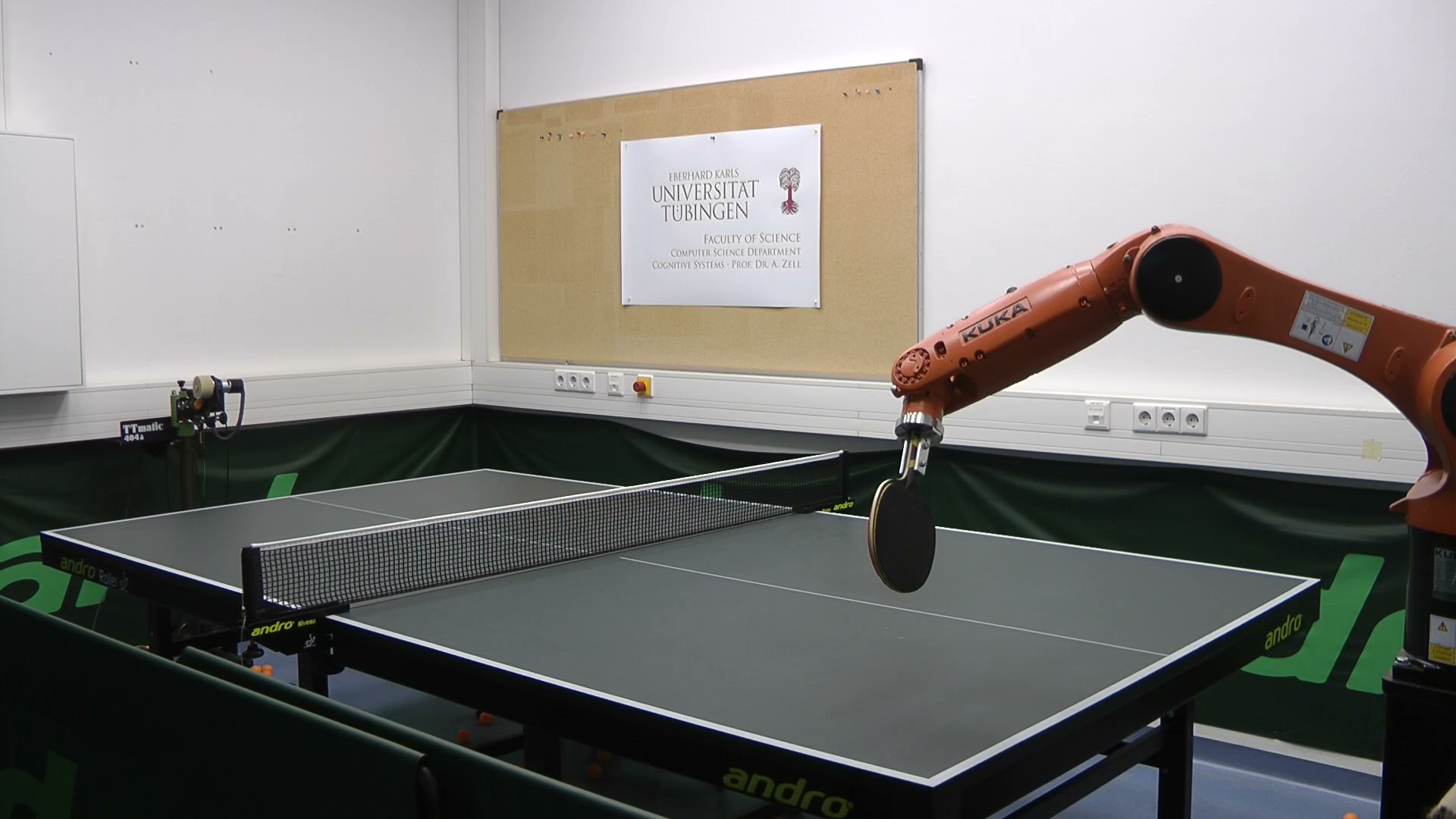}
	\caption{Testing in three scenarios: Player1, Player2, and Machine.}
	\label{fig:three_scen}
\end{figure}

To test our algorithms comprehensively, we conduct the experiments in three scenarios (see Fig. \ref{fig:three_scen}) in which the complexity gradually increases.
In the first scenario, a human player (Player1) serves the ball with different starting positions at the front of the table. In this way, the the hitting position along the $y$-axis can be completely covered.
In the second scenario, a human player (Player2) with higher skills than Player1 plays a long game rally to test the continuous performance of the robot. 
The state variables and the range of the ball for these two scenarios are shown in the third column of the Table~\ref{table:range_kuka}. 
In the last and most complex scenario, the ball throwing machine (Machine) is used to serve balls with various spins and speeds. 
Since many aspects such as the robot, the racket, the state of the ball, the human players as well as the evaluation metrics are completely different, it is difficult to fairly compare the performance of our algorithm with other previous works. Therefore, we report the Table~\ref{table:eval_reality} by directly using the data in~\cite{mulling2013learning, Kyohei2019ThePP} or by manually analyzing from~\cite{buchler2020learning, tebbe2020sample}.

\begin{table}[h]
	\caption{Testing in reality.}
	\begin{center}
		\begin{tabular}{l|cccc}
			\hline
			Scenarios  & episodes & $\epsilon_d$  & $\epsilon_h$  & success rate \\
			\hline\hline
			Büchler et al.~\cite{buchler2020learning}  & 107 &  76.9cm&- & 75\%\\
			\hline
			Mülling et al.~\cite{mulling2013learning} & 30 &  46.0cm&- & 97\%\\
			\hline
			Kyohei et al.~\cite{Kyohei2019ThePP}  & 100 &  22.5cm&-&99\%\\
			\hline
			Jonas et al.~\cite{tebbe2020sample} & 300 & 24.2cm&-&93.6\%\\
			\hline\hline
			Player1 &   40 & 20.3cm & 22.2cm &  \multirow{3}*{98\%}\\
			Player2 &   40 & 25.6cm & 23.5cm\\
			Machine &   40 & 28.8cm & 20.2cm\\
			\hline
		\end{tabular}
	\end{center}
	\label{table:eval_reality}
\end{table}
 
The average distance error $\epsilon_d$ and height error $\epsilon_h$ in the testing in three real scenarios are $24.9\pm9.0$ cm and $21.9\pm4.6$ cm, respectively.
Playing performance including some failure cases can be found at \url{https://youtu.be/SNnqtGLmX4Y}.

\section{Conclusions and future work}
\label{sec:typestyle}
In this work, we developed a realistic simulation for a table tennis robot. To learn the optimal stroke movement for the robot, we proposed a new policy gradient approach with TD3 backbone. Different algorithms were fairly evaluated in simulation using 1000 balls with a wide range of spins and speeds. To cross the domain from simulation to reality, a retraining approach was employed for the original racket and a coefficient-unknown racket racket. The test experiments showed a successful return rate of 98\% in three complicated scenarios. The total training time is about 1.5 hours, which means that our algorithm is very efficient for application in robotic table tennis. Instead of a constant target position on the table, one can simply train a higher-level policy with a random target to make the game more challenging for human players. Moreover, our approach can be easily adapted to other robots playing racket-based sports, such as tennis, badminton, or squash.

Although we have shown significant improvements in robotic table tennis in both simulation and reality, the current control approach and the mechanical structure of the robot still limits its application in the real world. For example, the robot will fail to return the ball if the incoming ball is too high or too slow, since the target cannot be reached at a fast enough speed. Also, our robot will not have sufficient reaction time if the ball is too fast (e.g. $\textgreater $10m/s) because the minimum communication time between the controller and the robot is 5 ms. In the future we plan to optimize the Reflexxes motion libraries to produce a more applicable trajectory for back spin balls. Instead of constraining the hitting position $p_x^b$ along the x-axis, we will parameterize it as one action to be learned in RL. For a more offensive stroke, we could also try to learn the angular velocities of the racket so that the robot can initiatively generate a spin ball. 

\section{Acknowledge}
We acknowledge the support of the Vector Stiftung and the KUKA Robotics Corporation.

\bibliography{sn-bibliography}


\end{document}